\documentclass[letterpaper, 10 pt, conference]{ieeeconf}
\IEEEoverridecommandlockouts  
\let\labelindent\relax
\usepackage{enumitem}

\usepackage{mathtools,amsthm}

\usepackage{graphics} 
\usepackage{epsfig}
\usepackage{bm}
\usepackage{times} 
\usepackage{amsmath,amssymb,amsfonts}
\usepackage{algorithmic}
\usepackage{graphicx}
\usepackage{subcaption}
\usepackage{textcomp}
\usepackage{varioref}
\usepackage{import}
\usepackage{framed,xcolor}
\usepackage{color}
\usepackage{comment}
\usepackage{multirow}
\usepackage{epstopdf}   
\usepackage{psfrag}
\usepackage{breqn}
\usepackage{float}
\usepackage{mathtools}
\usepackage{booktabs}
\usepackage{soul}
\usepackage{amsbsy}
\usepackage[bottom]{footmisc}
\usepackage{lineno}
\usepackage{cleveref}
\usepackage{microtype}
\usepackage{comment}

\graphicspath{{Figures/}}

\usepackage{graphicx}
\usepackage{tabularx,booktabs}
\usepackage{color}
\usepackage{multicol}
\usepackage{amsmath,amssymb}
\usepackage{amsfonts}
\usepackage{textcomp}
\usepackage{psfrag}
\usepackage{multimedia}
\usepackage{fancybox}
\usepackage{comment}
\usepackage{soul}

\definecolor{seb}{rgb}{0.8,1,0.8}
\definecolor{arash}{rgb}{0.8,0.8,1}

\newcounter{lastnote}

\title{Safe Human-to-Humanoid Motion Imitation Using Control Barrier Functions} 
\author{Wenqi Cai$^{1}$, John Abanes$^{2}$, Nikolaos Evangeliou$^{1}$, and Anthony Tzes$^{1,3}$
\thanks{$^{1}$The authors are with the Electrical Engineering Program, New York University Abu Dhabi (NYUAD), 129188, UAE. email: {wenqi.cai@nyu.edu; nikolaos.evangeliou@nyu.edu; anthony.tzes@nyu.edu}}
\thanks{$^{2}$The author is with the Electrical \& Computer Engineering Department, New York University, Brooklyn, NY 11201, USA. email: {john.abanes@nyu.edu}}
\thanks{$^{3}$This work was partially supported by the NYUAD Center for Artificial Intelligence and Robotics (CAIR), funded by Tamkeen under the NYUAD Research Institute Award CG010.}
}
\begin{document}
\bstctlcite{IEEEexample:BSTcontrol}
\maketitle
\thispagestyle{empty}
\pagestyle{empty}
\begin{abstract}
Ensuring operational safety is critical for human-to-humanoid motion imitation. This paper presents a vision-based framework that enables a humanoid robot to imitate human movements while avoiding collisions. Human skeletal keypoints are captured by a single camera and converted into joint angles for motion retargeting. Safety is enforced through a Control Barrier Function (CBF) layer formulated as a Quadratic Program (QP), which filters imitation commands to prevent both self-collisions and human-robot collisions. Simulation results validate the effectiveness of the proposed framework for real-time collision-aware motion imitation.
\end{abstract}
%%%%%%%%%%%%%%%%%%%
\section{Introduction}
Human-to-humanoid motion imitation is receiving increasing attention for intuitive human-robot interaction, teleoperation, and demonstration-based control. Conventional approaches often rely on wearable devices or dedicated motion-capture systems, which can provide accurate measurements but are costly, cumbersome, and less practical for lightweight daily deployment \cite{gonccalves2024leader}. In contrast, recent progress in camera-based human pose estimation enables a more accessible alternative, allowing human motion to be inferred directly from visual observations in real time \cite{boldo2024real,ramasubramanian2024evaluation}.

However, perception and retargeting alone are not sufficient for safe humanoid imitation. During real-time upper-body motion tracking, the robot may encounter both self-collision and human-robot collision, especially when the arms move in close proximity to the torso or toward the demonstrator. This makes safety a central challenge: how can one preserve responsive imitation while enforcing collision avoidance online? CBFs provide an effective tool for safety-critical control \cite{dai2023safe}, and related ideas have been explored in vision-aided collaborative robotic settings \cite{chaikalis2026vision}. Nevertheless, their use in vision-driven human-to-humanoid imitation with online robot-human geometric safety filtering remains largely unexplored.

\begin{figure*}[htbp!]
  \centering
  \includegraphics[width=0.9\linewidth]{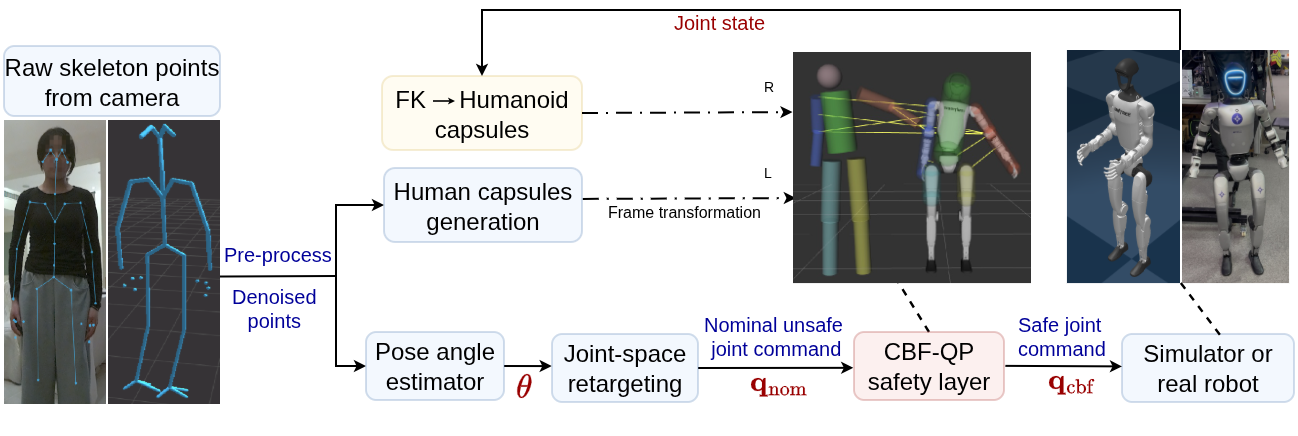}
  \caption{Overview of the proposed safe human-to-humanoid motion imitation framework.}
  \label{fig:framework}
\end{figure*}

Motivated by this gap, a safe human-to-humanoid imitation framework, shown in Fig.~\ref{fig:framework}, is developed. Starting from camera-observed skeleton keypoints, the system estimates a reduced human motion representation and maps it to the robot joint space through affine retargeting with joint-limit projection. A velocity-level capsule-based CBF layer then filters the nominal command to enforce both self-collision avoidance and robot-human collision avoidance, after which the safe command is executed in a high-fidelity simulator or on the real robot. In the presented implementation, imitation is restricted to the upper body, while the lower body is assumed to remain regulated by the built-in balance controller of the robot.

\textbf{Contributions.} 1) A vision-based safety framework is established for human-to-humanoid imitation, enabling online collision-aware motion generation from camera-based human observations; 2) A capsule-based CBF-QP formulation is used to handle both self-collision and human-robot collision within a unified real-time control layer; 3) Comparative benchmarks across multiple collision primitives are reported to assess computational trade-offs and to motivate the capsule-based modeling choice adopted in this work. These results provide a practical reference for future research on safe human-humanoid imitation and interactive humanoid control. While this paper focuses on simulation results, the framework is designed to transfer naturally to hardware settings.
%%%%%%%%%%%%%%%%%%%%%%%%%%%%%%%%%% 
\section{Human-to-Humanoid Motion Retargeting}
%%%%%%%%%%%%
\subsection{Pose Angle Estimation}
\label{sec:human_angle_estimation}
A reduced human pose is estimated directly from visual skeletal keypoints, without solving a full inverse-kinematics problem. To improve robustness, the raw skeleton points are first processed by a point-level filter that combines an Exponential Moving Average (EMA) with jump rejection. The same filtered skeleton stream is then used by both the pose angle estimator and the human capsule generation module, as shown in Fig.~\ref{fig:framework}. Let
\begin{equation}
\begin{aligned}
\bm{\theta} = [ 
\theta^{\mathrm{torso}}_{\mathrm{roll}}, \theta^{\mathrm{torso}}_{\mathrm{pitch}}, &\theta^{L}_{\mathrm{sh,pitch}}, \theta^{L}_{\mathrm{sh,roll}}, \theta^{L}_{\mathrm{el}}, \\
&\qquad\theta^{R}_{\mathrm{sh,pitch}}, \theta^{R}_{\mathrm{sh,roll}}, \theta^{R}_{\mathrm{el}} 
]^{\mathrm{T}} \in \mathbb{R}^8
\end{aligned}
\end{equation}
denote the reduced human pose. The skeleton is expressed in the global reference frame of the camera \(\mathcal{F}_Z\). A torso frame \(\mathcal{F}_T\) is constructed online from the pelvis and bilateral shoulder keypoints, with origin at the pelvis. Its upward axis is defined from the pelvis to the shoulder center, the lateral axis is obtained from the shoulder line after orthogonalization, and the remaining axis is determined by a right-handed cross product. This yields an orthonormal body-centered frame for expressing arm motion relative to the torso.

Using \(\mathcal{F}_T\), the torso roll and pitch are computed from the deviation of the torso frame from the global vertical direction, while the shoulder pitch and roll are obtained from the upper-arm vectors expressed in \(\mathcal{F}_T\). The elbow pitch is computed from the angle between the upper-arm and forearm segments. The $8$-DoF vector \(\bm{\theta}\) is then estimated as:
\begingroup
\setlength{\jot}{8pt}
\begin{equation*}
\resizebox{0.99\linewidth}{!}{$
\begin{aligned}
&\theta^{\mathrm{torso}}_{\mathrm{roll}}
=
\operatorname{atan2}\!\Big(
-\,{}^{Z}\mathbf{e}_{T,y}^{T}
\big(
{}^{Z}\mathbf{k}\times{}^{Z}\mathbf{e}_{T,z}
\big),\;
{}^{Z}\mathbf{k}^{T}{}^{Z}\mathbf{e}_{T,z}
\Big), \\
&\theta^{\mathrm{torso}}_{\mathrm{pitch}}
=
\operatorname{atan2}\!\Big(
-\,{}^{Z}\mathbf{e}_{T,x}^{T}
\big(
{}^{Z}\mathbf{k}\times{}^{Z}\mathbf{e}_{T,z}
\big),\;
{}^{Z}\mathbf{k}^{T}{}^{Z}\mathbf{e}_{T,z}
\Big), \\
&\theta^{L}_{\mathrm{sh,pitch}}
=
\operatorname{atan2}\!\big(
u^{T}_{L,y}, -u^{T}_{L,z}
\big), ~
\theta^{R}_{\mathrm{sh,pitch}}
=
\operatorname{atan2}\!\big(
u^{T}_{R,y}, -u^{T}_{R,z}
\big), \\
&\theta^{L}_{\mathrm{sh,roll}}
=
\operatorname{atan2}\!\big(
-\,u^{T}_{L,x}, -u^{T}_{L,z}
\big), ~
\theta^{R}_{\mathrm{sh,roll}}
=
\operatorname{atan2}\!\big(
u^{T}_{R,x}, -u^{T}_{R,z}
\big), \\
&\theta^{L}_{\mathrm{el}}
=
\arccos\!\left(
\frac{
{}^{Z}\mathbf{u}_{L}^{T}{}^{Z}\mathbf{f}_{L}
}{
\|{}^{Z}\mathbf{u}_{L}\|\,\|{}^{Z}\mathbf{f}_{L}\|
}
\right), \;
\theta^{R}_{\mathrm{el}}
=
\arccos\!\left(
\frac{
{}^{Z}\mathbf{u}_{R}^{T}{}^{Z}\mathbf{f}_{R}
}{
\|{}^{Z}\mathbf{u}_{R}\|\,\|{}^{Z}\mathbf{f}_{R}\|
}
\right),
\end{aligned}
$}
\end{equation*}
\endgroup
where \({}^{Z}\mathbf{k}=[0\;0\;1]^{T}\) is the global upward unit vector, \({}^{Z}\mathbf{e}_{T,x}\), \({}^{Z}\mathbf{e}_{T,y}\), and \({}^{Z}\mathbf{e}_{T,z}\) are the torso-frame axes expressed in \(\mathcal{F}_Z\); \({}^{Z}\mathbf{u}_{L}\), \({}^{Z}\mathbf{u}_{R}\), \({}^{Z}\mathbf{f}_{L}\), and \({}^{Z}\mathbf{f}_{R}\) denote the left/right upper-arm and forearm vectors expressed in \(\mathcal{F}_Z\), while \(u^{T}_{L,x}\), \(u^{T}_{L,z}\), \(u^{T}_{R,x}\), and \(u^{T}_{R,z}\) are the corresponding components of the left/right upper-arm vectors in \(\mathcal{F}_T\). 

\subsection{Joint-Space Retargeting}
\label{sec:retargeting}

The robot upper-body imitation space is chosen as
\[
\mathbf{q}
=
\begin{bmatrix}
q_{\mathrm{wr}} &
q_{\mathrm{wp}} &
q^{L}_{\mathrm{sp}} &
q^{L}_{\mathrm{sr}} &
q^{L}_{\mathrm{e}} &
q^{R}_{\mathrm{sp}} &
q^{R}_{\mathrm{sr}} &
q^{R}_{\mathrm{e}}
\end{bmatrix}^{T}
\in\mathbb{R}^{8},
\]
where \(q_{\mathrm{wr}}\) and \(q_{\mathrm{wp}}\) denote waist roll and waist pitch, and the remaining entries denote left/right shoulder pitch, shoulder roll, and elbow pitch.

Let \(\bm{\theta}_{\mathrm{home}}\) denote the calibrated neutral human pose and \(\mathbf{q}_{\mathrm{home}}\) denote the robot home configuration. The nominal retargeting map is defined as
\begin{equation}
\mathbf{q}_{\mathrm{nom}}
=
\Pi_{[\mathbf{q}_{\min},\,\mathbf{q}_{\max}]}
\!\left(
\mathbf{q}_{\mathrm{home}}
+
\mathbf{S}\bigl(\bm{\theta}-\bm{\theta}_{\mathrm{home}}\bigr)
+
\mathbf{b}
\right),
\label{eq:retarget_affine}
\end{equation}
where \(\mathbf{S}\in\mathbb{R}^{8}\) is a scaling vector, \(\mathbf{b}\in\mathbb{R}^{8}\) is a constant offset, and \(\Pi_{[\mathbf{q}_{\min},\,\mathbf{q}_{\max}]}\) denotes element-wise clipping to the admissible joint range. The resulting \(\mathbf{q}_{\mathrm{nom}}\) serves as the nominal imitation command before safety filtering and is forwarded to the CBF module as the unsafe upper-body joint command.

%%%%%%%%%%%%%%%%%%%%%%%%%%%%%%%%%%%%%%%%
\section{CBF-QP Safety Control}
\label{sec:cbf}
%%%%%%%%%%%%%%%%%%%%%%%%%%%%%%%%
\subsection{Capsule-Based Collision Model}
To represent collision geometry compactly, both the robot and the human are approximated by a corresponding seven-capsule model
\begin{equation*}
\mathcal{B}_{\{\mathrm{robot},\mathrm{human}\}}
=
\left\{
\begin{aligned}
&\texttt{torso, } \texttt{L/R upper arm,} \\
&\texttt{L/R forearm, } \texttt{L/R thigh}
\end{aligned}
\right\}.
\end{equation*}
Each capsule is parameterized by two endpoints and a radius. For the robot, these geometric parameters are updated from the current joint state through Forward Kinematics (FK). For the human, the capsules are extracted directly from the denoised skeleton points. The human capsules are further transformed into the robot reference frame before being used by the CBF safety layer.

%%%%%%%%%%%%%%%%%%%%%%%%%%%%%%%%%%%%%%%%%%%%%
\subsection{Collision-Avoidance CBF}
Consider a capsule pair \(A\) and \(B\), with radii \(r_A\) and \(r_B\), endpoints \((\mathbf{a}_A,\mathbf{b}_A)\) and \((\mathbf{a}_B,\mathbf{b}_B)\), and closest points \(\mathbf{p}_A\) and \(\mathbf{p}_B\). The closest-point distance is defined as
\begin{equation}
d_{AB}\big(\mathbf{x}_{AB}(\mathbf{q})\big)=\|\mathbf{p}_A-\mathbf{p}_B\|_2 ,
\label{eq:distance_ab}
\end{equation}
where \(\mathbf{x}_{AB}\) collects the endpoint positions needed to evaluate the pairwise proximity. For robot-attached capsules, \(\mathbf{x}_{AB}\) is determined by the robot joint configuration \(\mathbf{q}\) through forward kinematics. The corresponding safety function is defined as
\begin{equation}
h_{AB}(\mathbf{x}_{AB})
=
d_{AB}(\mathbf{x}_{AB})-(r_A+r_B+\phi),
\label{eq:capsule_barrier}
\end{equation}
where \(\phi>0\) is the barrier margin. Therefore, \(h_{AB}>0\) indicates safe separation, \(h_{AB}=0\) corresponds to the safety boundary, and \(h_{AB}<0\) indicates collision.

The standard first-order CBF condition is
\begin{equation}
\dot{h}_{AB}+\gamma h_{AB}\ge 0,
\label{eq:cbf_basic}
\end{equation}
with class-\(\mathcal{K}\) gain \(\gamma>0\). Let the control input be the joint velocity $\mathbf{u}:=\dot{\mathbf{q}}$, then, by the chain rule,
\begin{equation}
\dot{h}_{AB}
=
\frac{\partial h_{AB}}{\partial \mathbf{x}_{AB}}
\frac{\partial \mathbf{x}_{AB}}{\partial \mathbf{q}}
\mathbf{u}=\mathbf{A}_{AB}\mathbf{u},
\label{eq:h_dot_general}
\end{equation}
where \(\partial \mathbf{x}_{AB}/\partial \mathbf{q}\) is obtained from the kinematic Jacobians of the corresponding capsule endpoints. For self-collision, \(\mathbf{A}_{AB}\) is determined by the kinematics of both capsules. For robot-human collision, the human capsule is treated as fixed over one control step, and \(\mathbf{A}_{AB}\) is therefore computed using only the robot-side kinematics.

Accordingly, each active pair \(i\) yields the linear CBF constraint
\begin{equation}
\dot{h}_i(\mathbf{q},\mathbf{u})
\ge
-\gamma h_i(\mathbf{q})
\quad\text{i.e.,}\quad
\mathbf{A}_i(\mathbf{q})\,\mathbf{u}
\ge
-\gamma h_i(\mathbf{q}).
\label{eq:cbf_linear_constraint}
\end{equation}
This constraint regulates the evolution of the closest-point distance \(d_i\) between the capsule pair. When \(h_i>0\), \(d_i\) is allowed to decrease, so that the motion can remain close to the nominal imitation behavior. As \(h_i\to 0\), further decrease is restricted to avoid crossing the safety boundary. If \(h_i<0\), the constraint enforces an increase in \(d_i\), driving the pair back toward safe separation.
%%%%%%%%%%%%%%%%%%%%%%%%%%%%%%
\subsection{Real-Time CBF-QP}
The retargeting stage provides a nominal joint-position command \(\mathbf{q}_{\mathrm{nom}}\in\mathbb{R}^{8}\). The safety layer maintains an internal safe target \(\mathbf{q}_{\mathrm{cbf}}\in\mathbb{R}^{8}\), from which the reference joint velocity $\mathbf{u}_{\mathrm{ref}}$ is generated as
\begin{equation}
\mathbf{u}_{\mathrm{ref}}
=
K\left(\mathbf{q}_{\mathrm{nom}}-\mathbf{q}_{\mathrm{cbf}}\right),
\label{eq:u_ref}
\end{equation}
where \(K>0\) is a proportional gain. The safe joint velocity \(\mathbf{u}^{\star}\) is obtained by solving the QP
\begin{equation}
\begin{aligned}
\arg\min_{\mathbf{u}}&\quad
\frac{1}{2}
\left\|
\mathbf{u}-\mathbf{u}_{\mathrm{ref}}
\right\|_{W}^{2}
\\
\text{s.t.}\quad
&
\mathbf{A}_i(\mathbf{q})\,\mathbf{u}
\ge
-\gamma h_i(\mathbf{q}),
\quad i=1,\dots,N_c,
\\
&
\mathbf{u}_{\min}\le \mathbf{u}\le \mathbf{u}_{\max},
\end{aligned}
\label{eq:cbf_qp}
\end{equation}
where \(W\succ0\) weights the deviation from the nominal reference and \(N_c\) is the number of active CBF constraints. The internal safe target is then updated by
\begin{equation}
\mathbf{q}_{\mathrm{cbf}}^{+}
=
\mathbf{q}_{\mathrm{cbf}}
+
\Delta t\,\mathbf{u}^{\star},
\label{eq:q_cbf_update}
\end{equation}
and the updated \(\mathbf{q}_{\mathrm{cbf}}^{+}\) is passed to the downstream position controller. This formulation preserves the nominal imitation behavior when no collision constraints are active, while smoothly modifying the motion whenever self-collision or human-robot collision constraints become binding.
%%%%%%%%%%%%%%%%%%%%%%%%%%%%%%%%%%%%%%%%%%%%%%%%%%%%%%%%%%%%%%%%
%%%%%%%%%%%%%%%%%%%%%%%%%%%%%%%%%%%%%%%%%%%%%%%%%%%%%%%%%%%%%%%
\section{Simulation Studies}
\label{sec:results}
The proposed framework is evaluated from two complementary perspectives: 1) Closed-loop safety performance in representative self-collision and human-humanoid interaction scenarios, evaluated by comparing executions with and without the CBF-QP; 2) Computational benchmarking of different collision-geometry models used in the CBF formulation, together with efficiency and constraint-complexity comparisons that support the capsule-based model used in this work.
% The simulation pipeline follows the proposed method:
% visual (ZED camera) skeleton
% $\rightarrow$ human angle estimation
% $\rightarrow$ joint-space retargeting
% $\rightarrow$ unsafe command
% $\rightarrow$ CBF-QP
% $\rightarrow$ safe command
% $\rightarrow$ MuJoCo execution.
%%%%%%%%%%%%%%%%%%%%%%%%%%%%%
\subsection{Safety in Motion Imitation}
\begin{figure}[htbp!]
    \centering
    \begin{subfigure}{\linewidth}
        \centering
        \includegraphics[width=0.5\linewidth]{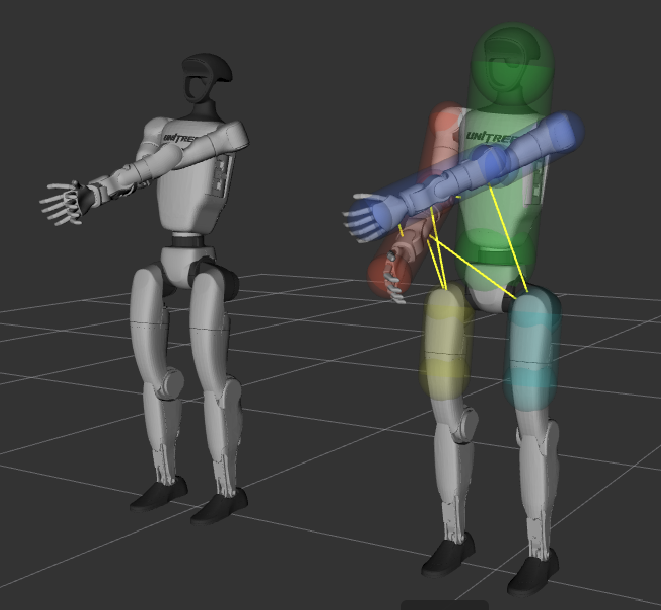}
        \caption{Representative scenario: cross-arm reach. The ghost robot (left) shows unsafe arm overlap, whereas the CBF-filtered robot (right) maintains a safety margin. The yellow segments indicate selected monitored CBF constraint pairs.}
        \label{fig:self_snap}
    \end{subfigure}
    \begin{subfigure}{\linewidth}
        \centering
        \includegraphics[width=\linewidth]{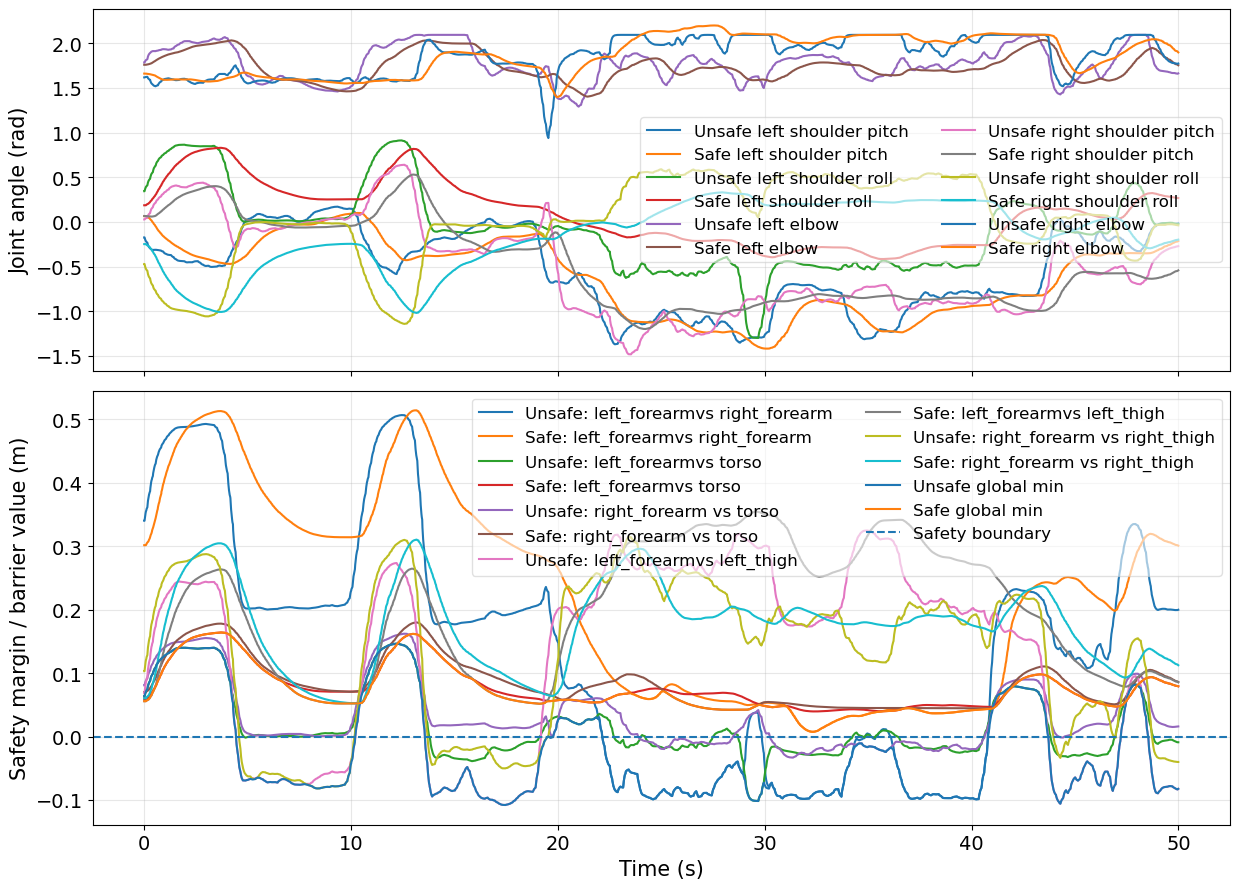}
        \caption{Representative joint trajectories (top) and safety margins (bottom) together with the global minimum over all monitored constraints.}
        \label{fig:self_curve}
    \end{subfigure}
    \caption{Self-collision avoidance with and without CBF-QP.}
    \label{fig:self_results}
\end{figure}

\begin{figure}[htbp!]
    \centering
    \begin{subfigure}{\linewidth}
        \centering
        \includegraphics[width=0.5\linewidth]{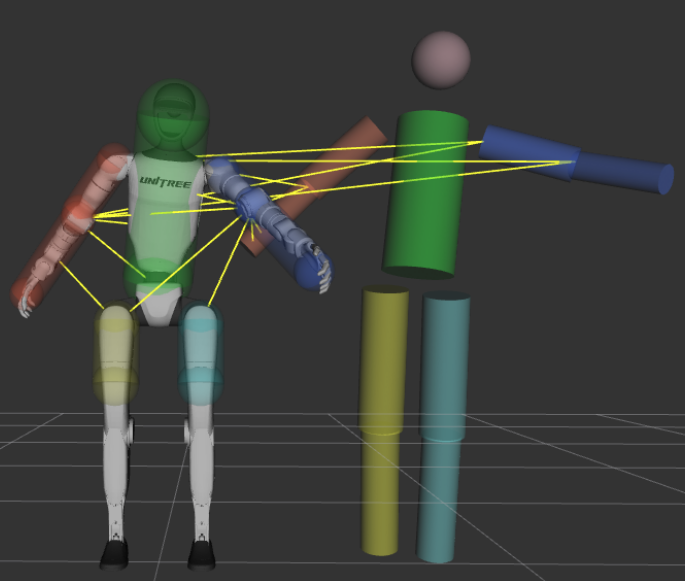}
        \caption{Representative scenario: side-by-side arm raise. The yellow segments indicate selected monitored CBF constraint pairs.}
        \label{fig:hh_snap}
    \end{subfigure}
    \begin{subfigure}{\linewidth}
        \centering
        \includegraphics[width=\linewidth]{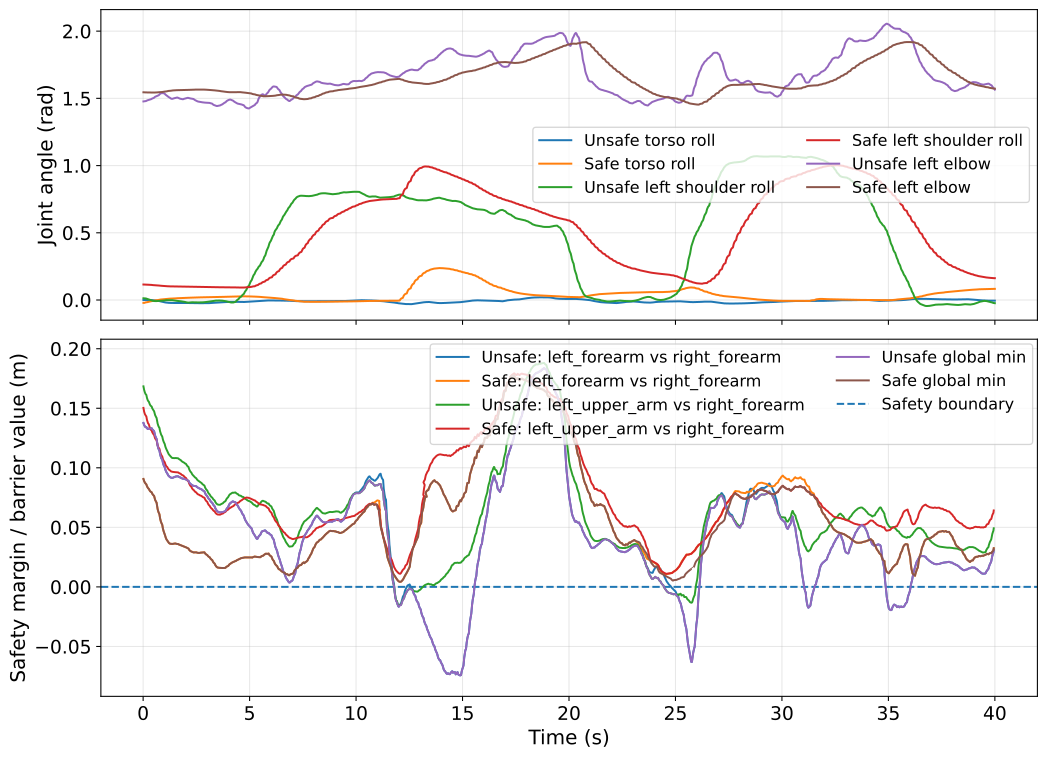}
        \caption{Representative joint trajectories (top) and safety margins (bottom) together with the global minimum over all monitored constraints.}
        \label{fig:hh_curve}
    \end{subfigure}
    \caption{Human-humanoid collision avoidance with and without CBF-QP.}
    \label{fig:hh_results}
\end{figure}

We investigate two typical safety-critical scenarios: a self-collision case (\texttt{cross-arm reach}) and a human-humanoid collision case (\texttt{side-by-side arm raise}). It is observed that the nominal retargeted motion remains qualitatively consistent with the demonstrated human motion, while the CBF layer intervenes only when collision constraints become active. The detailed results are shown in Fig.~\ref{fig:self_results} and~\ref{fig:hh_results}. The yellow lines in the scenario snapshots indicate selected monitored CBF constraint pairs through their closest-point connections. For visual clarity, only a subset of representative pairs is shown. In both cases, the nominal unsafe motion violates the safety boundary, whereas the CBF-QP effectively preserves safety margins while retaining meaningful imitation behavior.

%%%%%%%%%%%%%%%%%%%%%%%%%%%%%%%%
%%%%%%%%%%%%%%%%%%%%%%%%%%%%%%%%
\subsection{CBF Geometry Benchmarks}
\begin{figure}[htbp!]
    \centering
    \includegraphics[width=0.99\linewidth]{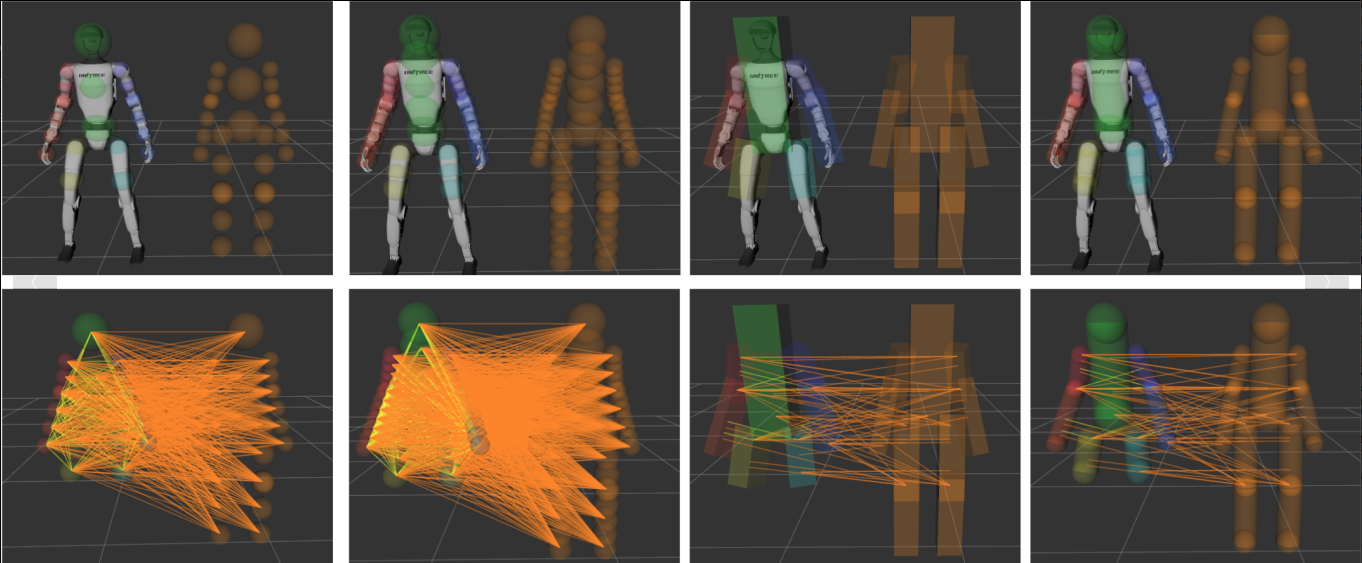}
    \caption{Benchmarked collision geometry. From left to right: sphere decomposition with no subsampling, sphere decomposition with one subsampling, bounding box decomposition, capsule decomposition. Lines represent CBF constraints.}
    \label{fig:collider_methods}
\end{figure}

%%%%%%%%%%%%%%%%%%%%%%%%
To motivate the use of capsule collision geometry, performance is compared with other commonly used models, including spheres and bounding boxes, in the human-robot simulation environment. Table~\ref{tab:benchmarks} shows the performance results of various collider geometries. For sphere decomposition, the center-to-center squared distance is used, implemented via~\cite{morton2025oscbf}. For capsules, the signed distance between sphere-swept volumes method is used and implemented via~\cite{tracy_diffpills_2022}. For boxes, the convex polytope proximity method is used, implemented via~\cite{tracy_differentiable_2023}. These methods all provide differential proximity measurements that CBF can utilize. Benchmarks are performed on a computer with an Intel Core Ultra 9 275HX CPU and an NVIDIA RTX 5080 Laptop GPU. A visualization of these collision models is shown in Fig.~\ref{fig:collider_methods}.

%%%%%%%%%%%%%%%%%%%
%%%%%%%%%%%%%

The computational performance of different implementation backends is also benchmarked. As shown in Table~\ref{tab:benchmarks}, utilizing the GPU results in significant performance uplift due to the number of parallel constraints. Differentiable kinematics are implemented in the CPU via Pinocchio~\cite{carpentier2019pinocchio} while GPU methods utilize JAX~\cite{jax2018github}. The best overall performance was demonstrated by capsule geometry on the GPU, demonstrating both low solve time and minimal constraint blow-up. 
%%%%%%%%%%%%%%
\begin{table}[htbp!]
\centering
\caption{CBF Benchmark Statistics for Various Collision Geometries}
\label{tab:benchmarks}
\setlength{\tabcolsep}{3pt}
\renewcommand{\arraystretch}{1.1}
\scriptsize
\begin{tabular}{@{}l*{4}{r}@{}}
\toprule
& \multicolumn{4}{c}{\textbf{CPU}} \\
\cmidrule(l){2-5}
Metric & Spheres\,(k\!=\!0) & Spheres\,(k\!=\!1) & Boxes & Capsule \\
\midrule
Rate (Hz) & 10.02 & 8.5 & 10.424 & 4.113 \\
\# Constraints & 705 & 1988 & \textbf{72} & \textbf{72} \\
Initial Compile Time (s) & \textbf{0.02} & 0.05 & 0.4 & 0.7 \\
Solve Time (s) & 0.100 & 0.118 & 0.096 & 0.243 \\
\midrule
& \multicolumn{4}{c}{\textbf{GPU}} \\
\cmidrule(l){2-5}
Metric & Spheres\,(k\!=\!0) & Spheres\,(k\!=\!1) & Boxes & Capsule \\
\midrule
Rate (Hz) & 30.5 & 11.549 & 3.937 & \textbf{33.084} \\
\# Constraints & 705 & 1988 & \textbf{72} & \textbf{72} \\
Initial Compile Time (s) & 29.4 & 184.8 & 36.7 & 29.4 \\
Solve Time (s) & 0.033 & 0.087 & 0.254 & \textbf{0.030} \\
\bottomrule
\end{tabular}
\vspace{2pt}\\
\raggedright
\scriptsize\textit{Note:} $k$ denotes the degree of spherical subsampling used to approximate capsule geometry. All benchmarks use a QP solver with a maximum of $100$ iterations and a solution tolerance of $10^{-3}$. Results are reported for a worst-case setting in which CBF constraints are imposed between all external colliders and all other internal colliders, highlighting computational differences across geometry models.
\end{table}
%%%%%%%%%%%%%%%%%%%%%%%%%%%%%%%%%%%%%%%%%%%
%%%%%%%%%%%%%%%%%%%%%%%%%%%%%%%%%%%%%%%%%%%%
\section{Conclusion}
\label{sec:conclusion}
This late-breaking result introduced a vision-based framework for safe human-to-humanoid motion imitation, unifying pose angle estimation, retargeting, and capsule-based CBF-QP safety filtering. The presented framework shows that real-time collision-aware imitation can be achieved within a unified control architecture. Future work will focus on real-robot implementation, quantitative evaluation of imitation fidelity, and more efficient CBF formulations with richer collision pair sets.
%%%%%%%%%%%%%%%%%%%%%
% \begin{appendices}
% \section{Parameters of the dynamics and RL}\label{app:1}
% \begin{table}[H]
% \renewcommand\arraystretch{1.4}
% \centering \scriptsize

% \begin{tabular}{c|c|c}
% \hline
% %\small
% \multirow{5}*{Dynamics} &$\phi_b$ & 5 \\ \cline{2-3}
%  &$\phi_s$ & 2.5 \\ \cline{2-3} 
%  &$\alpha$ & 1/12\\ \cline{2-3}
%   &$\bar U$ & 1 \\ \cline{2-3}
%  &$\Delta$ & $\mathcal N\left(0, 0.05\right)$\\\hline
%  \multirow{2}*{RL} &$\Phi$ & $\left[\left( {s - 0.5} \right)^2,s,1\right]^\top$ \\\cline{2-3} 
%  & $p$ & 1000\\\cline{2-3} 
%  \hline
% \end{tabular}
% \end{table} 
% \end{appendices}

\bibliographystyle{IEEEtran}
\bibliography{reference}
\end{document}